\pdfoutput=1

\documentclass[11pt]{article}

\usepackage[]{acl}

\usepackage{times}
\usepackage{latexsym}
\usepackage{graphicx}
\usepackage{multirow}
\usepackage{subfigure}

\usepackage[T1]{fontenc}

\usepackage[utf8]{inputenc}

\usepackage{microtype}

\title{ChronosLex: Time-aware Incremental Training for Temporal Generalization of Legal Classification Tasks}

\author{Santosh T.Y.S.S \and Tuan-Quang Vuong \and Matthias Grabmair \\
School of Computation, Information, and Technology \\ Technical University of Munich, Germany \\
\texttt{\{santosh.tokala, quang.vuong, matthias.grabmair\}@tum.de}}

\begin{document}
\maketitle
\begin{abstract}
This study investigates the challenges posed by the dynamic nature of legal multi-label text classification tasks, where legal concepts evolve over time. Existing models often overlook the temporal dimension in their training process, leading to suboptimal performance of those models over time, as they treat training data as a single homogeneous block. To address this, we introduce ChronosLex, an incremental training paradigm that trains models on chronological splits, preserving the temporal order of the data. 
However, this incremental approach raises concerns about overfitting to recent data, prompting an assessment of mitigation strategies using continual learning and temporal invariant methods. Our experimental results over six legal multi-label text classification datasets reveal that continual learning methods prove effective in preventing overfitting thereby enhancing temporal generalizability, while temporal invariant methods struggle to capture these 
dynamics of temporal shifts.  

\end{abstract}

\section{Introduction}
Legal classification tasks involve the assignment of legal documents, texts, or cases to specific legal categories, making them essential for legal practitioners, researchers, and organizations seeking efficient legal document management and analysis. However, the dynamic nature of legal domain, characterized by the intricate interplay of laws, precedents, and ever-evolving interpretation of existing jurisprudence, influenced by external forces such as real-world events and shifting societal norms, provides a distinctive challenge for these legal tasks specifically. In this backdrop of a dynamic and non-stationary world, temporal generalization of models emerges as a key necessity, given such systems are deployed to assist users with data that it encounters in the future, whilst being trained on data from the past.

\begin{figure}
\centering 
\scalebox{0.4}{
 \includegraphics{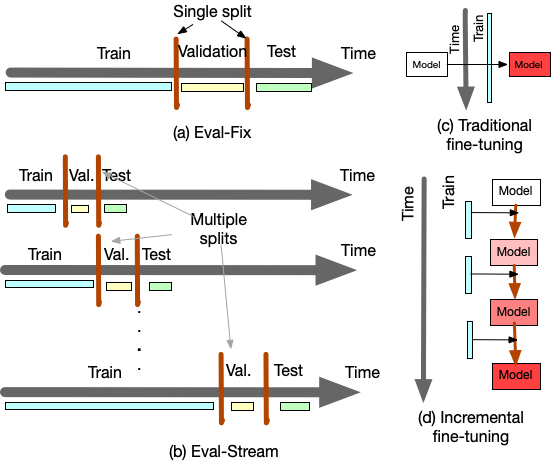}}
    \caption{Different evaluation and training strategies employed in this work, to assess and improve temporal generalization of models for legal classification tasks.}
    \label{fig:overview}
\end{figure}

Most of the works dealing with legal classification often neglect time as a factor and simply split the data randomly into training, validation, and test sets (e.g., \citealt{paul2020automatic,malik2021ildc,zheng2021does,tuggener2020ledgar,hendrycks2021cuad,papaloukas2021multi,lippi2019claudette, puaiș2021named,luz2018lener,csulea2017predicting} inter alia). Recent works have realized that this standard splitting criterion leads to overly optimistic performance estimates, which are far from realistic scenarios when such systems need to be deployed to assist with the new content over time and suffer from degradation of performance due to temporal misalignment i.e., difference in distributions of training and evaluation data instances. They advocate for chronological splitting of datasets to estimate the performance, accounting for the data drift over time (e.g., \citealt{chalkidis2021paragraph, chalkidis2022lexglue,chalkidis2021multieurlex,xu2023vechr,niklaus2021swiss,chalkidis2022improved} inter alia). Furthermore, we posit that using a single fixed chronological test split (Eval-Fix, Fig. \ref{fig:overview}a) 
may yield misleading results if the chosen fixed split coincides with anomalous events, such as Brexit, Covid-19, lockdown, the Russo-Ukrainian War, or political regime changes, impacting one side of the splits more than the other. To address this concern, we utilize a streaming evaluation protocol (Eval-Stream, Fig. \ref{fig:overview}b) which involves evaluating the model over multiple splits involving distinct time periods which can lead to reliable conclusions about models' robustness. 
In this protocol, we incorporate all past data into training and assess the model's capacity to adapt to emerging trends in the next time period, which mirrors standard machine learning development pipelines, where models are regularly updated and redeployed to account for temporal drifts \cite{yao2022wild}.

While legal texts are intrinsically tied to the time of their origin, making it imperative to account for these temporal aspects when developing classification models, these previous methods overlook the chronological context in which legal documents exist during the training process and consider the entire training data spanning across a considerable stretch of time, as a single homogeneous block (Fig. \ref{fig:overview}c). This oversight hinders model capabilities to adapt to the data drifts that naturally occur over time leading to performance degradation when evaluated over data from the future time \cite{gaspers2022temporal, rijhwani2020temporally}. 

For instance, shifts in societal attitudes towards fundamental rights like privacy, discrimination, and freedom of expression drive the evolution of legal concepts, broadening or contracting their scope over time. Influenced by technological advances and global events, national legal concepts expand to encompass emerging trends, as seen in security related matters to encompass artificial intelligence and cryptocurrency. Landmark cases can also introduce new civil/human rights, such as the recognition of environmental rights within the right to life. Legislative changes, such as developments in data protection laws keeping pace with technology, further shape concepts, reflecting evolving regulatory needs. Overall, data drifts in the legal domain can manifest gradually through the gradual evolution of legal concepts 
, reflecting societal changes, emerging legal trends, and real-world events or even abruptly through landmark legal decisions, overruling precedents, and legislative changes.

In response to these dynamic temporal drifts, we put forth a hypothesis which we subsequently validate: Models trained on data from a timeframe closer to the test set tend to yield superior results, under the assumption of the same model and dataset size. Motivated by this finding, and with the aim to capitalize on the 
whole training set — typically associated with better performance with more data, we introduce ``ChronosLex'', an incremental training framework (Fig. \ref{fig:overview}d). In this approach, the model is systematically trained on data from one time period at a time before progressing to the next chronological split, thus preserving the temporal order of the data the model interacts with during training. Our experiments on six multi-label legal classification datasets demonstrate that incremental training paradigm improves temporal generalization of these models on three datasets but also leads to overfitting to recent data on other datasets.

To further improve the models' robustness to address the overfitting issue, we assess (i) different families of continual learning algorithms such as regularization-based EWC \cite{kirkpatrick2017overcoming}, rehearsal-based ER \cite{chaudhry2019tiny}, AGEM \cite{lopez2017gradient}, and parameter expansion based Adapters \cite{houlsby2019parameter}, LoRA \cite{hu2021lora} (ii) temporally invariant methods \cite{yao2022wild} derived from domain adaptation methods such as GroupDRO \cite{sagawa2019distributionally}, DeepCORAL \cite{sun2016deep}, IRM \cite{arjovsky2019invariant} where we treat consecutive windows of time as distinct domains. Our experimental results suggest that temporal invariant methods fail to learn these  dynamics while continual learning methods effectively prevent models from overfitting to recent data, ultimately enhancing temporal generalization. 

\section{Related Work}
\noindent \textbf{Temporal Drift} There has been increasing recent evidence that temporal misalignment results in degradation of model performance due to difference in distribution between the train and test data, caused due to temporal shifts \cite{jaidka2018diachronic,yao2022wild,gorman2019we}. These distribution shifts arise with the passage of time and are a very naturally occurring type of distribution shift due to non-stationarity and evolving world \cite{schlimmer1986incremental,widmer1993effective}. Temporal misalignment can be caused due to (1) the dynamic nature of language \cite{rosin2022time,rottger2021temporal,loureiro2022timelms,agarwal2022temporal,amba2021dynamic,rijhwani2020temporally,luu2022time,jaidka2018diachronic} and (2) the update of factual information \cite{margatina2023dynamic,jang2021towards,jang2022temporalwiki,lazaridou2021mind,dhingra2022time,liska2022streamingqa}. 

Temporal generalization has been explored both in upstream Language Model pre-training \cite{lazaridou2021mind,loureiro2022timelms,jang2021towards,jang2022temporalwiki,dhingra2022time,jin2022lifelong,amba2021dynamic
} and in downstream tasks, such as sentiment analysis \cite{lukes2018sentiment,agarwal2022temporal,guo2023predict}, named entity recognition \cite{rijhwani2020temporally,onoe2022entity}, question answering \cite{liska2022streamingqa,shang2022improving}, headline generation \cite{sogaard2021we}, rumor/fake news detection \cite{mu2023s,hu2023learn}, spoken language understanding \cite{gaspers2022temporal}. model explainability \cite{zhao2022impact}, document classification 
\cite{rottger2021temporal,chalkidis2022improved,huang2018examining}, abusive language detection \cite{jin2023examining,florio2020time}, topic modeling \cite{zhang2023vibe} and readmission prediction in the health care context \cite{guo2022evaluation,guo2023ehr}. 

In this work, we assess temporal generalization in multi-label legal text classification, which is challenging given the complexities of managing both the multiplicity of labels and accommodating the evolving legal context over time. This is in contrast to prior studies that predominantly focused on single-label classification. In prior work on temporal generalization in multi-label legal text classification, \citealt{chalkidis2022improved} postulated that temporal drift is primarily attributable to shifts in the label distribution over time and explored various group-robust optimization algorithms to mitigate disparities at the label level. We propose an alternative viewpoint, arguing that temporal drift in legal tasks extends beyond label distribution shifts to encompass shifts in the input text as well, driven by the fact that the vocabulary changes with time (Tab. \ref{distribution}). Furthermore, even vocabulary specific to each label undergoes temporal shifts (Tab \ref{distribution}). This complexity was not fully considered in \citealt{chalkidis2022improved}, where the entire training data was treated as a single unit, overlooking shifts in input text distribution over time. To address this, we introduce an incremental training framework 
which enables models to adapt to distribution shifts over time. We further improve its robustness 
by evaluating a range of continual learning and temporal invariant approaches for multi-label legal text classification.


\noindent \textbf{Continual Learning} Most research in continual learning (or lifelong learning) initially centered around computer vision tasks, and more recently, it has gained attention in the NLP field \cite{ke2021adapting,ke2021continual,biesialska2020continual,ke2022continual,sun2019lamol,wang2019sentence}. The majority of these works adopted traditional task-incremental, domain-incremental, or label-incremental settings. While our temporal incremental setting bears resemblance to the domain-incremental setting, a crucial distinction lies in the assumption of strict boundaries in domain incremental settings, which is not applicable to our temporal adaptation where the boundaries between drifts are blurred \cite{prabhu2020gdumb,aljundi2019gradient}. Generally, continual learning algorithms can be categorized into   (i) Rehearsal-based methods \cite{rolnick2019experience,rebuffi2017icarl}, which maintain a memory buffer of older data to perform experience replay with actual data \cite{de2019episodic}, automatically generated data \cite{sun2019lamol}, or previously computed gradients \cite{lopez2017gradient},
(ii) Regularization-based approaches \cite{kirkpatrick2017overcoming,chen2020recall,huang2021continual}, which regularizes neural network parameters from drastic updates for new information to preserve the information of older ones, preventing overfitting to the newer data (iii) Parameter-expansion methods \cite{qin2022elle,gururangan2022demix,yoon2018lifelong} which freeze network architectures on older data and dynamically grow branches for newer ones. These methods have been explored in the pre-training stage to accommodate factual updates \cite{jang2021towards,jang2022temporalwiki} or adaptation to new domains \cite{jin2022lifelong,gururangan2022demix} and in the fine-tuning stage \cite{yao2022wild,lin2022continual}, which is closely related to our work, but within the context of single-label classification tasks. To the best of our knowledge, continual methods have never been explored for multi-label legal classification tasks.

\noindent \textbf{Temporal invariant Domain Adaptation} Domain adaptation (DA) primarily addresses the co-variate shift between source and target data distributions \cite{ruder2019neural}. This involves learning domain-invariant feature representations that can generalize across different domains \cite{bollegala2011relation,ganin2016domain}. In the legal context, DA methods have been applied to tasks such as Legal Judgement Prediction, where reasoning with respect to each article is treated as a domain \cite{tyss2023zero} and legal text classification tasks, where each label is considered a domain to address label imbalance \cite{chalkidis2022improved}. Traditional DA methods are typically applied to domains with clear distinctions, whereas temporal distribution shifts tend to occur gradually without well-defined boundaries, forming a continuous process. To adapt domain invariant methods to address temporal distribution shifts, we assume instances from consecutive windows of time to constitute a distinct domain, enabling the application of invariant learning approaches to these constructed domains \cite{yao2022wild}. To the best of our knowledge, this is the first attempt to assess these methods for multi-label legal texts to handle temporal drift.

\section{Evaluation Strategy}
\label{eval}
To assess temporal generalizability, following previous works \cite{sogaard2021we,chalkidis2022lexglue}, we carry out evaluation using fixed chronological splits, referred to as \textbf{Eval-Fix} (Fig. \ref{fig:overview}a). In this setup, we split the entire data chronologically into train ($d_{< t_1}$), validation ($d_{\geq t_1 \& \leq t_2}$), and test ($d_{> t_2}$) splits, where $d_t$ denotes data from time period $t$, and $t_1$ < $t_2$. We train and validate the model using the respective splits and subsequently report its performance over the entire test set. 

We posit that such a fixed split may not reflect a true picture as it can be influenced by the interplay between the splits and the data distribution where the anomalous/drift-driven events may overlap or primarily fall within one of the splits, leading to misleading conclusions about the model's ability to handle temporal drifts. To mitigate the impact of dataset-specific traits on the splits, we adopt a streaming evaluation protocol, inspired by \citealt{yao2022wild} and \citealt{lopez2017gradient}, which we refer to as \textbf{Eval-Stream} (Fig. \ref{fig:overview}b). We evaluate models using multiple splits, formed at distinct timestamps, considering all the data up to time period $t$ ($d_{\leq t}$) for training, $d_t$ for validation, and $d_{t+1}$ as the test split. We focus only on the evaluation of the future data rather than the past because the practical need for performance over past data is limited. While it is possible to test the model on data from all subsequent time periods ($d_{>t}$) at each time period $t$, we restrict the testing to the following time period $t+1$, to simplify the result analysis, aligning with practical scenarios where models are updated based on their performance in the next time iteration.

\section{Datasets}
We conduct our experiments on the 
six multi-label legal classification datasets 
from three sources.

\noindent \textbf{UKLEX} \cite{chalkidis2022improved} This dataset comprises legislative documents of the UK, publicly accessible through the National Archives, which are typically categorized into thematic areas such as healthcare, finance, education, transportation and planning, which are outlined in the document preamble and serve as indexes for archival purposes. The dataset contains 36.5k documents and is chronologically split into training (20k, 1975–2002), validation (8.5k, 2002–2008), and test (8.5k, 2009–2018) sets. The labels are provided at two distinct levels of granularity, encompassing 18 and 69 topics, referred to as Small (S) and Medium (M) respectively. For \emph{eval-fix}, we use the above splits and for \emph{eval-stream}, we consider a two-year time period as a unit. Due to limited data prior to 1990 and to ensure an adequate number of data instances for model learning, we report eval-stream performance from 1992. 




\noindent \textbf{EURLEX} \cite{chalkidis2021multieurlex} This dataset contains EU legislation documents accessible via the EUR-Lex platform and is annotated using concepts from EuroVoc, a thesaurus maintained by the EU's Publications Office. We work with the English portion of this dataset, consisting of 65k documents and split chronologically into training (55k, 1958–2010), validation (5k, 2010–2012), and test (5k, 2012–2015) sets which we use for \emph{eval-fix}. The dataset provides four levels of label granularity. We use the first two levels of the EuroVoc taxonomy, as followed in \citealt{chalkidis2022improved}, encompassing 21 and 127 concepts, denoted as Small (S) and Medium (M), respectively. For \emph{eval-stream}, we use two years as one unit and report performance from 1987 due to a lower number of instances in the initial years.

\noindent \textbf{ECHR} The dataset by \cite{chalkidis2019neural,chalkidis2021paragraph} comprises cases heard by the European Court of Human Rights, which are publicly accessible via HUDOC, the official court database. These cases involve the adjudication of complaints by individuals against states for alleged violations of their rights as enshrined in the European Convention of Human Rights. Each case includes information about the convention articles that have been alleged to be violated and which the court has found to be violated. The dentification of alleged and violated articles are referred to as Task B and A, respectively, by \citealt{chalkidis2022lexglue}. It consists of 11K cases, divided chronologically into training (9k, 2001–2016), validation (1k, 2016–2017), and test (1k, 2017–2019) sets for \emph{eval-fix}. 
We use the 14 and 17 articles related to the core rights as followed in \citealt{valvoda2023role} for Task A and B respectively. We use annual data as a unit for \emph{eval-stream} and report the performance from 2004.

While task employed in this work are all effectively text classification, the tasks surveyed here are different in nature. The ECHR tasks represent legal determinations by judges in specific cases of high complexity and semantic depth, whereas the keyword classification tasks of EURLEX and UKLEX represent semantically much shallower analysis. This has consequences for the types and complexity of temporal shifts one would observe. For example, the court changing its jurisprudence on a particular legal issue prompted by world events is a more complex phenomenon than the usage of particular European Law Database Keyword.

\section{ChronosLex}
Previous approaches for legal classification typically fine-tune models on the entire training dataset in a shuffled manner, treating all data as a single homogeneous block. We argue that such an approach neglects the temporal nature of the data which may be crucial to learn the temporal evolution of concepts. To address this limitation, we propose the ChronosLex framework, which considers the chronological context of data to capture and adapt models to the natural drifts over time. We hypothesize that recent data holds insights into upcoming distributional shifts and training models with this recent data enables better adaptation to temporal changes. Thus, we systematically train the model with data from one time period at a time before progressing to the next chronological split. Specifically, at time $t$, the model $m_t$ is initialized with the model obtained from the previous timestamp $m_{t-1}$ and fine-tuned on data $d_t$ from time period $t$, progressively moving to the next time period $t+1$. We term this approach \textit{Incremental Fine-tuning (\textbf{IFT})} (Fig. \ref{fig:overview}d), where the model architecture and the loss function remain similar to traditional fine-tuning, but the model encounters chronological training data incrementally.  

However, IFT may risk overfitting to recent data, potentially forgetting previously acquired knowledge which is crucial for extrapolating to the new distribution based on past data. Therefore, a balance between preserving knowledge from the past and accumulating information from recent data is essential to enhance temporal generalization. To explore this balance, we investigate continual learning methods and temporal invariant methods, 
to adapt to temporally emerging distribution shifts.

\subsection{Continual Learning Methods}
The aim of continual learning is to accumulate knowledge incrementally without forgetting information from previous steps referred to as catastrophic forgetting. In our specific context, we explore the efficacy of these methods in enabling models to extrapolate into the future based on past information, within the framework of a boundary-unaware, non-stationary temporal shift setting.

\noindent \textbf{EWC} \cite{kirkpatrick2017overcoming} Elastic Weight Consolidation is a regularization-based method which adds a temporal regularization term to the task-specific actual loss so that the parameter changes from $t-1$ to $t$ are restricted to avoid over-fitting. It produces a weighted penalty such that the parameters that are more important to the previous time stamp will have larger penalty weights, 
to balance the trade-off between previous knowledge and new knowledge. 
It uses Fisher Information Matrices to estimate the importance of parameters to use them as the weighted penalty.

\noindent \textbf{ER} \cite{rolnick2019experience} Experience Replay falls into the category of rehearsal-based methods that stores samples from previous time stamps into a growing memory module. We use instances from memory as additional training examples along with current time stamp instances. 

\noindent \textbf{A-GEM} \cite{lopez2017gradient} Average-Gradient Episodic Memory, a rehearsal method, leverages an episodic memory to store a sample of examples from previous time stamp similar to ER, but additionally equip the loss on older samples as inequality constraints, avoiding their increase. 

\noindent \textbf{LoRA} \cite{hu2021lora} falls into the category of parameter-expansion method which freezes the original parameters of the pre-trained model and introduces trainable low-rank matrices and combines them with the original matrices in the multi-head attention and are updated during fine-tuning.

\noindent \textbf{Adapters} \cite{houlsby2019parameter} is another parameter expansion method that freezes the original parameters of the pre-trained model and injects two small modules between the self-attention sub-layer and the feed-forward sub-layer inside each layer of transformer sequentially. The adapter module consists of a down-projection, an up-projection, and a nonlinear function between them with a residual connection across each module.

\begin{table}
\centering
\scalebox{0.9}{
\begin{tabular}{|l|cc|cc|}
\hline
          & \multicolumn{2}{c|}{$x$}                                               & \multicolumn{2}{c|}{$x|y$}           \\ \hline
& \multicolumn{1}{c|}{Old} & \multicolumn{1}{c|}{Rec.} & \multicolumn{1}{c|}{Old} & \multicolumn{1}{c|}{Rec.} \\ \hline
UKLEX(S)   & \multicolumn{1}{c|}{\multirow{2}{*}{0.314}} & \multirow{2}{*}{0.264} & \multicolumn{1}{c|}{0.384} & 0.328 \\ \cline{1-1} \cline{4-5} 
UKLEX(M)   & \multicolumn{1}{c|}{}                       &                        & \multicolumn{1}{c|}{0.412} & 0.366 \\ \hline
EURLEX(S)  & \multicolumn{1}{c|}{\multirow{2}{*}{0.359}} & \multirow{2}{*}{0.265} & \multicolumn{1}{c|}{0.379} & 	0.288
 \\ \cline{1-1} \cline{4-5} 
EURLEX(M) & \multicolumn{1}{c|}{}                       &                        & \multicolumn{1}{c|}{0.409} & 0.324\\ \hline
ECHR(A)     & \multicolumn{1}{c|}{\multirow{2}{*}{0.236}} & \multirow{2}{*}{0.155} & \multicolumn{1}{c|}{0.306} & 	0.247\\ \cline{1-1} \cline{4-5} 
ECHR(B)     & \multicolumn{1}{c|}{}                       &                        & \multicolumn{1}{c|}{0.318	} & 0.266  \\ \hline
\end{tabular}}
\caption{Jensen–Shannon divergence score between the split of training set (Old/Recent) and the test set over the vocabulary distribution ($x$) and vocabulary conditioned on the label ($x|y$). Higher Score indicates more divergence from the test set distribution.  }
\label{distribution}
\end{table}

\subsection{Temporal Invariant Methods}
Domain invariant representation learning boosts model transferability across domains by eliminating domain-specific information, preventing overfitting to specific domains, and improving generalization to unseen target domains. In our case, we aim to build a temporally invariant model by excluding features tied to specific time periods. The difficulty of their application in our context lies in the lack of well-defined boundaries for temporal distribution shifts, with no explicit signal for drift awareness. 
To tackle this, 
we treat every sliding window of length L timestamps as a domain. 

\noindent \textbf{DeepCORAL} \cite{sun2016deep} 
Correlation Alignment penalizes the differences in the mean and covariance of the feature distributions of each domain to obtain domain invariant representations.

\noindent \textbf{IRM} \cite{arjovsky2019invariant} Invariant Risk Minimization aims to penalize variance across multiple training dummy estimators across domains, i.e., performance cannot vary across samples corresponding to the same domain. 

\noindent \textbf{GroupDRO} \cite{sagawa2019distributionally} Group Distributionally Robust Optimization aims to optimize the worst-domain loss wherein the domain-wise losses are weighted inversely proportional to the performance of instances in that domain.

\section{Experiments}
\subsection{Base Model \& Metrics}
For UKLEX and EURLEX, we use a state-of-the-art model, BERT-LWAN \cite{chalkidis2020empirical}, which has a label-wise attention network on top of the pre-trained model. Specifically, we pass the text into LegalBERT \cite{chalkidis2020legal} and employ one attention head per label to generate N document representations where N denotes the number of labels, which are finally passed through a linear layer to get the final predictions. For ECHR, we employ a state-of-the-art hierarchical version of BERT due to long input documents \cite{chalkidis2022lexglue}. Specifically, we pass each sentence in the long document into LegalBERT to obtain [CLS] representations which are then passed through a two-layer transformer to obtain final representations. Implementation details of all the methods can be found in Appendix \ref{app-implementation}. Following \citealt{chalkidis2022improved}, we report, micro-F1, macro-F1, and mean R-Precision (m-RP).

\begin{table*}[!ht]
\centering
\scalebox{0.8}{
\begin{tabular}{|l|c|c|c|c|c|c|c|c|c|}
\hline
                & \multicolumn{3}{c|}{ULKEX(S)} & \multicolumn{3}{c|}{ULKEX(M)} & \multicolumn{3}{c|}{EURLEX(S)} \\
                \hline
Method          & mac.-F1  & mic.-F1  & mRP    & mac.-F1  & mic.-F1  & mRP    & mac.-F1   & mic.-F1  & mRP    \\ \hline
Baseline - Old  & $72.92_{3.86}$ & $78.26_{3.55}$ & $76.02_{4.13}$ & $50.54_{6.73}$ & $67.34_{1.36}$ & $62.72_{2.37}$  & $59.16_{0.66}$ & $72.19_{0.99}$ & $66.53_{0.60}$ \\
Baseline - Rec. & $73.26_{1.67}$ & $79.34_{2.47}$ & $77.20_{2.83}$ & $52.38_{8.96}$ & $69.84_{1.33}$ & $66.01_{1.42}$  & $64.08_{0.36}$ & $76.20_{0.85}$ & $71.62_{0.27}$ \\ \hline
Baseline - Full & ${74.96}_{3.44}$ & $80.61_{2.14}$ & $78.66_{1.66}$ & $54.89_{12.34}$ & $70.85_{2.42}$ & $67.25_{0.55}$ & $67.70_{1.01}$ & $78.57_{0.97}$ & $73.46_{0.28}$ \\ \hline
IFT             & $78.65_{2.92}$ & $82.29_{1.91}$ & $80.20_{1.78}$ & $55.18_{11.79}$ & $72.57_{1.45}$ & $69.34_{2.65}$ & $66.28_{0.91}$ & $77.71_{1.23}$ & $72.48_{0.63}$ \\ \hline
EWC             & $80.15_{1.74}$ & $83.80_{2.17}$ & $82.41_{2.37}$ & $55.66_{8.91}$ & $74.79_{2.30}$ & $\underline{72.22}_{3.81}$  & $68.01_{0.66}$ & $78.69_{0.70}$ & $73.69_{0.20}$ \\
ER              & $\underline{80.26}_{1.77}$ & $\underline{83.82}_{2.83}$ & $\underline{83.08}_{4.20}$ & $\underline{55.93}_{9.58}$ & $\underline{75.03}_{3.22}$ & $71.99_{3.40}$  & $\mathbf{68.27}_{0.68}$ & $\underline{78.84}_{0.83}$ & $\underline{74.02}_{0.43}$ \\
AGEM            & $79.13_{1.69}$ & $83.25_{2.17}$ & $81.39_{2.44}$ & $55.32_{9.44}$ & $73.84_{2.84}$ & $70.81_{4.03}$  & $67.80_{0.86}$ & $78.60_{0.82}$ & $73.27_{0.34}$ \\
LORA            & $80.03_{1.95}$ & $83.67_{2.78}$ & $82.58_{2.79}$ & $55.77_{9.95}$ & $74.54_{3.89}$ & $71.15_{4.45}$  & $\underline{68.05}_{0.11}$ & $\mathbf{79.02}_{0.87}$ & $\mathbf{74.14}_{0.44}$ \\
Adapter         & $\mathbf{80.64}_{0.67}$ & $\mathbf{84.15}_{3.99}$ & $\mathbf{83.13}_{4.11}$ & $\mathbf{56.27}_{10.84}$ & $\mathbf{75.67}_{4.00}$ & $\mathbf{72.65}_{4.29}$ & $67.96_{0.68}$ & $78.80_{1.17}$ & $73.57_{0.65}$ \\ \hline
DeepCORAL       & ${75.51}_{3.66}$ & ${80.40}_{2.54}$ & ${77.67}_{1.52}$ & ${50.58}_{8.26}$ & $70.25_{2.46}$ & $65.03_{2.39}$  & $64.71_{1.93}$ & ${74.98}_{2.21}$ & ${70.05}_{1.81}$ \\
IRM             & $77.82_{4.73}$ & $80.69_{0.88}$ & $78.98_{0.65}$ & $52.11_{10.08}$ & $70.29_{0.40}$ & ${64.15}_{0.04}$ & ${62.47}_{2.42}$ & $75.46_{2.77}$ & $70.07_{2.99}$ \\
GroupDRO        & $77.11_{3.76}$ & $80.73_{1.11}$ & $78.76_{0.77}$ & $51.45_{10.09}$ & ${70.06}_{1.16}$ & $64.98_{1.43}$ & $63.57_{0.68}$ & $76.91_{0.96}$ & $71.40_{0.70}$ \\

\hline
\hline

                  & \multicolumn{3}{c|}{EURLEX(M)} & \multicolumn{3}{c|}{ECHR(A)} & \multicolumn{3}{c|}{ECHR(B)} \\ \hline
Baseline - Old  & $38.35_{0.81}$ & $59.32_{1.90}$ & $52.23_{1.51}$  & $47.72_{3.05}$ & $62.86_{2.41}$ & $59.09_{3.06}$  & $43.77_{2.48}$ & $70.54_{1.50}$ & $64.25_{0.68}$ \\
Baseline - Rec. & $43.69_{1.26}$ & $64.64_{1.80}$ & $56.36_{1.59}$  & $52.31_{5.76}$ & $64.52_{0.20}$ & $60.14_{0.87}$  & $49.11_{3.99}$ & $74.96_{1.32}$ & $70.26_{2.05}$ \\ \hline
Baseline - Full & $44.17_{1.25}$ & $66.84_{2.02}$ & $58.66_{1.13}$  & $53.31_{8.36}$ & $66.95_{0.61}$ & $62.68_{1.03}$  & $\mathbf{51.93}_{1.07}$ & $\mathbf{75.27}_{0.18}$ & $71.89_{1.39}$ \\ \hline
IFT             & $44.97_{2.23}$ & $67.64_{1.88}$ & $59.08_{2.16}$  & $52.44_{7.11}$ & ${65.87}_{0.84}$ & ${61.55}_{0.97}$  & $\mathbf{50.82}_{0.79}$ & $\mathbf{75.13}_{2.43}$ & $71.09_{1.49}$ \\ \hline
EWC             & $44.67_{2.40}$ & $67.49_{2.28}$ & $58.88_{1.55}$  & $\mathbf{56.25}_{10.33}$ & $67.58_{0.29}$ & $62.43_{1.30}$ & $50.01_{3.68}$ & $74.86_{0.41}$ & $70.21_{0.61}$ \\
ER              & $45.42_{1.58}$ & $67.88_{1.99}$ & $59.72_{0.98}$  & $\underline{55.87}_{7.76}$ & $\underline{68.55}_{1.75}$ & $63.99_{0.15}$  & $49.31_{2.51}$ & $74.84_{0.04}$ & $70.51_{0.29}$ \\
AGEM            & $44.48_{1.93}$ & $66.77_{1.81}$ & $58.42_{1.18}$  & $54.93_{8.39}$ & $67.26_{0.18}$ & $63.66_{0.10}$  & $49.87_{3.19}$ & $\underline{75.11}_{0.61}$ & $\underline{73.51}_{1.47}$ \\
LORA            & $\mathbf{47.43}_{1.76}$ & $\mathbf{68.75}_{1.99}$ & $\underline{60.18}_{1.21}$  & $53.76_{5.89}$ & $66.82_{0.38}$ & $62.94_{1.01}$  & $50.31_{2.31}$ & $74.37_{0.72}$ & $72.71_{2.05}$ \\
Adapter         & $\underline{46.15}_{1.34}$ & $\underline{68.57}_{1.52}$ & $\mathbf{60.83}_{1.24}$  & $52.84_{3.76}$ & ${66.30}_{1.24}$ & ${62.13}_{0.69}$  & $\underline{50.59}_{1.15}$ & $74.79_{0.16}$ & $\mathbf{73.57}_{1.97}$ \\ \hline
DeepCORAL       & ${39.57}_{1.73}$ & ${63.07}_{1.69}$ & ${55.41}_{1.73}$  & ${50.48}_{5.98}$ & $68.19_{1.76}$ & $\underline{64.25}_{1.59}$  & ${48.57}_{2.29}$ & ${70.77}_{2.41}$ & ${68.94}_{2.13}$ \\
IRM             & $40.66_{1.05}$ & $65.12_{1.45}$ & $57.55_{1.15}$  & $51.62_{2.91}$ & $\mathbf{68.57}_{0.77}$ & $\mathbf{64.26}_{3.69}$  & $49.96_{1.60}$ & $73.87_{0.11}$ & $70.19_{1.48}$ \\
GroupDRO        & $40.11_{2.33}$ & $64.86_{2.02}$ & $57.99_{1.84}$  & $51.35_{0.13}$ & $68.32_{3.62}$ & $63.16_{2.68}$  & $49.24_{2.64}$ & $71.35_{1.53}$ & $69.97_{1.55}$

\\ \hline
\end{tabular}}
\caption{Performance of different categories of methods on the Eval-Fix setting over six datasets. Best and Second best values in each metric is bolded and underlined respectively. Subscript refers to standard deviations.}
\label{eval-fix}
\end{table*} 

\subsection{Results on Eval-Fix Setting} We present the results on eval-fix in Table \ref{eval-fix}. 

\noindent \textbf{Quantifying Temporal Shift:} To understand the effect of temporal distance between the training and test data, we divide the training data into two versions: (i) the first half of chronologically sorted training data referred to as \emph{Old}, and (ii) the latter half referred to as \emph{Recent}. To quantify the temporal distribution shifts between these splits and the test set, we calculate the Jensen-Shannon divergence score between the distribution of the vocabulary ($x$). 
There is a possibility that the observed vocabulary distribution shift might be influenced by changes in label distribution over time. We further disentangle it by calculating conditional vocabulary distribution for each label ($x|y$) and report the average of divergence scores across all labels. This specifically assesses how the vocabulary associated with each label changes over time. From Table \ref{distribution}, we observe that the recent split, which is temporally closer to the test data, has a lower divergence score compared to older split, confirming our hypothesis of temporal distribution drift over time. It is noteworthy that this may underestimate the effect, as we are specifically calculating the drift at the lexical level without capturing semantic shifts over time (changes in associated meaning or contextual usage of specific words).

To further elucidate the impact of this temporal drift on model performance, we evaluate the model on the eval-fix test set by training the baseline model using \emph{Old} and \emph{Recent} splits of training set. To remove a possible confounding effect of dataset size, each of these two models has access to the same number of training instances. As shown in Table \ref{eval-fix}, \emph{baseline-Recent} consistently outperforms \emph{baseline-Old} across all metrics and datasets. This result validates our hypothesis that \textbf{models trained on temporally closer data to the test set tend to yield superior results, under the assumption of the same model and dataset size} as we observed that temporally closer data deviates less from the test set in vocabulary distribution. Finally, we use the whole training data to create \emph{Baseline-Full} model and it exhibits enhanced performance across all metrics and datasets, underscoring the effectiveness of a larger dataset to develop a temporally robust generalizable classifier.

\noindent \textbf{Baseline vs. IFT} From Table \ref{eval-fix}, we observe \emph{IFT} performs better than \emph{Baseline-Full} in UKLEX(S,M) and EURLEX(M) but falls short in EURLEX(S) and ECHR(A,B). While \emph{baseline} employs a traditional fine-tuning approach ignoring the temporal order of training dataset, IFT preserves the chronological order of the input dataset by training the model incrementally. This in turn assists the model in detecting drifts over time and adapting to them accordingly, simultaneously harnessing the whole dataset. However, this strategy may lead to overfitting to recent data, as evidenced by the lower performance in EURLEX(S) and ECHR(A,B). Mitigating this requires a mechanism to revisit older data to preserve their information while accumulating and adapting to newer trends/drifts.

\noindent \textbf{Continual Learning Methods} On UKLEX(S), all continual learning methods exhibit superior performance compared to both IFT and baseline approaches. Notably, Adapters and ER stand out, surpassing others, with EWC and LoRA following closely behind. AGEM shows comparatively lesser efficacy among them. This trend of continual learning methods outperforming IFT persists in the UKLEX(M) scenario, although the margin of improvement on macro-F1 scores is relatively narrow, given the larger number of labels in the medium (M) split coupled with high label imbalance.

\begin{table*}[!ht]
\centering
\scalebox{0.79}{
\begin{tabular}{|l|c|c|c|c|c|c|c|c|c|}
\hline
                & \multicolumn{3}{c|}{ULKEX(S)} & \multicolumn{3}{c|}{ULKEX(M)} & \multicolumn{3}{c|}{EURLEX(S)} \\
                \hline
Method   & macro-F1  & micro-F1  & m-RP       & macro-F1  & micro-F1  & m-RP        & macro-F1  & micro-F1  & m-RP   \\ \hline
Baseline & $78.97_{4.33}$ & $83.89_{3.08}$ & $83.42_{3.60}$  & $55.60_{6.18}$ & $75.87_{3.43}$ & $71.52_{4.23}$  & $64.65_{6.35}$    & $78.18_{3.96}$   & $73.05_{5.40}$  \\ \hline
IFT      & $79.98_{4.55}$ & $84.91_{3.12}$ & $83.65_{3.61}$  & $56.15_{6.26}$ & $77.12_{3.48}$ & $73.03_{4.33}$  & $63.26_{4.78}$    & $77.99_{3.25}$   & $72.61_{4.60}$  \\ \hline
EWC      & $81.42_{5.21}$ & $86.12_{3.49}$ & $85.15_{4.26}$  & $58.08_{6.66}$ & $79.31_{5.85}$ & $75.23_{6.83}$  & $\underline{65.16}_{5.79}$    & $78.64_{3.74}$   & $73.38_{5.12}$  \\
ER       & $81.51_{4.76}$ & $85.91_{3.20}$ & $85.14_{3.87}$  & $58.17_{6.51}$ & $\underline{79.38}_{5.64}$ & $\mathbf{75.55}_{6.69}$  & $64.95_{6.03}$    & $78.69_{3.64}$   & $73.31_{5.08}$  \\
AGEM     & $\underline{81.78}_{5.41}$ & $86.11_{3.64}$ & $\underline{85.23}_{4.26}$  & $\mathbf{58.45}_{6.94}$ & $79.26_{5.64}$ & $75.22_{6.84}$  & $65.07_{6.02}$    & $78.67_{3.76}$   & $73.47_{5.34}$  \\
LORA     & $\mathbf{81.99}_{3.76}$ & $\mathbf{86.19}_{3.68}$ & $\mathbf{85.38}_{4.23}$  & $\underline{58.31}_{7.15}$ & $79.12_{5.77}$ & $75.03_{6.81}$  & $\mathbf{65.56}_{5.66}$    & $\mathbf{78.86}_{3.53}$   & $\mathbf{73.61}_{4.88}$  \\
Adapter  & $81.14_{5.46}$ & $\underline{86.16}_{3.66}$ & $85.17_{4.31}$  & $58.22_{7.24}$ & $\mathbf{79.72}_{5.45}$ & $\underline{75.54}_{6.47}$  & $64.82_{6.25}$    & $\underline{78.77}_{3.78}$   & $\underline{73.56}_{5.17}$  \\ \hline
DeepCORAL & ${77.51}_{3.61}$ & ${83.14}_{3.13}$ & ${81.60}_{4.20}$ & ${51.56}_{6.93}$ & $74.96_{4.26}$ & $69.85_{4.88}$  & ${57.31}_{2.34}$    & ${75.70}_{2.62}$   & ${69.43}_{3.85}$  \\
IRM      & $77.83_{4.42}$ & $83.48_{2.92}$ & $82.02_{3.27}$  & $52.94_{6.97}$ & $74.79_{4.07}$ & $70.45_{4.42}$  & $61.21_{3.86}$    & $76.68_{2.89}$   & $71.17_{4.42}$  \\
GroupDRO & $78.05_{4.71}$ & $83.56_{4.33}$ & $82.86_{4.53}$  & $52.70_{8.17}$ & ${74.30}_{5.58}$ & ${69.39}_{5.66}$  & $61.88_{3.13}$    & $76.75_{2.38}$   & $71.17_{3.49}$  \\
\hline
\hline

                  & \multicolumn{3}{c|}{EURLEX(M)} & \multicolumn{3}{c|}{ECHR(A)} & \multicolumn{3}{c|}{ECHR(B)} \\ \hline

Baseline & $35.59_{9.45}$ & $66.16_{6.16}$ & $58.99_{7.61} $ & $51.38_{6.78}$ & $69.30_{3.66}$ & $63.96_{3.33} $ & $47.22_{4.17}$ & $74.86_{2.08}$ & $67.56_{3.10}$ \\ \hline
IFT      & ${34.61}_{9.27}$ & $65.89_{6.11}$ & $58.15_{7.45} $ & $49.86_{6.52}$ & ${68.27}_{3.56}$ & ${63.08}_{3.34} $ & $46.73_{4.10}$ & $74.27_{2.07}$ & $67.09_{3.17}$ \\ \hline
EWC      & $38.65_{7.15}$ & $66.87_{6.33}$ & $59.81_{7.85}$ & $\mathbf{52.79}_{7.17}$ & $70.23_{4.47}$ & $65.05_{4.64} $ & $46.60_{5.39}$ & $75.01_{2.27}$ & $68.15_{3.32}$ \\
ER       & $\underline{39.61}_{7.27}$ & $66.88_{6.30}$ & $59.92_{7.66}$ & $51.82_{6.78}$ & $70.18_{4.16}$ & $64.98_{4.03} $ & $\underline{47.46}_{5.33}$ & $74.91_{2.42}$ & $\mathbf{68.44}_{3.29}$ \\
AGEM     & $38.27_{7.64}$ & $66.73_{6.33}$ & $59.57_{7.95}$ & $51.53_{6.74}$ & $69.86_{4.22}$ & $65.12_{3.94} $ & $46.70_{4.70}$ & $\underline{75.05}_{1.94}$ & $68.12_{2.76}$ \\
LORA     & $\mathbf{40.39}_{7.70}$ & $\underline{67.16}_{6.55}$ & $\mathbf{60.16}_{7.92}$ & $\underline{52.54}_{6.01}$ & $70.43_{4.15}$ & $65.74_{3.62} $ & $47.09_{5.16}$ & $\mathbf{75.13}_{2.65}$ & $\underline{68.39}_{2.93}$ \\
Adapter  & $38.57_{7.78}$ & $\mathbf{67.63}_{6.85}$ & $\underline{60.14}_{8.26}$ & $52.35_{6.73}$ & $70.04_{4.36}$ & $64.99_{4.20} $ & $\mathbf{47.64}_{5.60}$ & $74.78_{2.45}$ & $67.83_{3.26}$ \\ \hline
DeepCORAL & $30.50_{6.44}$ & ${60.62}_{5.75}$ & ${53.14}_{6.66}$ & $45.48_{3.65}$ & $72.19_{1.94}$ & $66.95_{2.34} $ & ${40.81}_{5.86}$ & ${70.27}_{3.14}$ & ${65.37}_{3.45}$ \\
IRM      & $34.82_{8.38}$ & $62.76_{5.47}$ & $55.36_{6.57} $ & $48.74_{4.61}$ & $\mathbf{76.32}_{2.71}$ & $\mathbf{69.72}_{3.21} $ & $45.82_{4.23}$ & $73.42_{2.34}$ & $68.33_{2.35}$ \\
GroupDRO & $34.69_{7.00}$ & $63.99_{5.37}$ & $56.36_{6.38} $ & ${46.02}_{5.50}$ & $\underline{74.75}_{5.11}$ & $\underline{68.75}_{4.90} $ & $45.30_{2.58}$ & $72.93_{3.92}$ & $66.68_{5.96}$ \\ \hline
\end{tabular}}
\caption{Aggregated performance of various method categories in the Eval-Stream setting across six datasets.
Best and Second best values in each metric is bolded and underlined respectively. Subscript refers to standard deviations.
}
\label{eval-stream}
\end{table*}

Transitioning to EURLEX(S), continual learning methods again outshine IFT and the baseline, with LoRA and ER taking the lead. As observed in UKLEX(S,M), AGEM falls short again. Surprisingly, Adapters, which excels on UKLEX(S,M), does not exhibit the same level of performance on EURLEX(S). On EURLEX(M), Adapters, LoRA, and ER outperform both the baseline and IFT, while the other methods remain comparable. AGEM, once again, demonstrates lower performance. 

Shifting focus to ECHR(A), EWC, ER, and AGEM outperform the baseline and IFT. Notably, Adapters exhibits the least favorable performance in this context. On ECHR(B), none of the continual learning methods succeed in surpassing IFT and baseline on micro- and macro-F1 scores. However, AGEM, LoRA, and Adapters outperform them on the m-RP metric. The underperformance on ECHR(B) can be attributed to the spurious correlations in the dataset, as highlighted in \citealt{santosh2022deconfounding}. These spurious attributes, inadvertently learned during training, persist over time, posing a challenge for the model to effectively unlearn them. Continual learning, while adept at adapting to evolving information, faces difficulties in fully addressing and mitigating these learned artifacts.

Among these methods, overall AGEM falls short in most of the datasets and we attribute it to its design of stricter inequality constraint giving the model less freedom to accumulate new knowledge.  

\noindent \textbf{Temporal Invariant Methods} On UKLEX(S), temporal invariant methods outperform baseline but fall short compared to IFT. However, they struggle to cross the baseline in UKLEX(M), EURLEX(S,M), and ECHR(B) datasets. While on ECHR(A), they outperform IFT and even continual methods on micro-F1 and m-RP but fall short on macro-F1  metric. We speculate that the objective of temporal invariant methods aims to eliminate the distribution by learning generalizable feature representation, while the adaptive objective for the continual methods helps effectively to track and adapt to the shifts. 


In summary, these findings collectively suggest that applying continual learning methods on top of incremental fine-tuning leads to improved performance in 5 out of 6 datasets. This underscores their capacity to adapt to natural shifts that occur over time, while also preserving past knowledge which assists them to extrapolate trends into the future from historical distribution.

\subsection{Results on Eval-Stream Setting} 
Under the Eval-Stream setting, we assess performance through multiple splits, conducting evaluations on each subsequent timestamp, as detailed in Sec. \ref{eval}. The performance visualization over the multiple splits for each method is presented in Appendix \ref{app-eval}. We report the averaged results across all the splits for each method in Table \ref{eval-stream}.

\noindent \textbf{IFT vs. Baseline} IFT performs better than baseline on UKLEX(S,M) only and underperforms on the other four datasets. This observed trend aligns with our eval-fix analysis, due to IFT's strong reliance on recent data, potentially leading to overfitting.

\noindent \textbf{Continual Learning Methods} All the continual Learning methods perform consistently better than IFT and baseline, over all the metrics across all the six datasets. This contrasts with our conclusions drawn in the eval-fix setting, where we did not observe such uniform performance improvement across all the methods. This discrepancy could be attributed to our reliance on conclusions drawn from a single test split in eval-fix, which may exhibit a distinct pattern of shift. These findings emphasize the importance of adopting a streaming evaluation protocol in datasets where distribution shifts occur naturally over time. Rather than relying solely on one split, which may lead to misleading conclusions about changes in performance and method efficacy, using streaming evaluation ensures more reliable and comprehensive insights.

Among the continual learning methods, LoRA demonstrates superior temporal robustness which can be attributed to its design of additional weight matrices. This design allows LoRA to effectively discern and assimilate additional drift compared to previous knowledge, accumulating it through additional parameters and thereby preserving older knowledge. Following closely, ER emerges as the next best performer, emphasizing the importance of revisiting past knowledge to guide the model away from heavy reliance on recent data. AGEM falls short and can be attributed to its stricter inequality constraint, as discussed before.

\noindent \textbf{Temporal Invariant Methods}
All these approaches underperform compared to the baseline. Among them, IRM performs comparably to baseline, followed closely by GroupDRO. Notably, DeepCORAL (DC) consistently lags behind across all datasets. This can be attributed to the rationale behind these approaches. While IRM and GroupDRO employ penalties on loss functions so that performance cannot vary across samples of different time periods, DC directly minimizes the difference between feature representations across samples of different time periods. This design causes models trained with DC to suppress distribution shifts rather than actively learn and adapt to recent drifts.

\section{Conclusion}

In this study, we delve into temporal drift in legal multi-label text classification tasks, revealing a crucial oversight in the current model training process that treats the training data as a single, homogeneous block. This neglect results in the degradation of performance over time. To remedy this, we introduce ChronosLex, an incremental training paradigm that allows the model to interact with data chronologically, facilitating adaptation to temporal shifts. However, we observe a potential overfitting to recent data with this incremental approach. 
To address overfitting, we explore mitigation strategies using both continual learning and temporal invariant approaches and found that Continual algorithms exhibit promising results by leveraging historical distribution to extrapolate trends into the future. In contrast, temporal invariant methods prove less effective in tackling this. Finally, we advocate for streaming evaluation protocol with multiple time splits to draw reliable conclusions, especially when time is involved as a critical axis. In future, we plan to evaluate this incremental strategy across other legal tasks such as prior case retrieval, contract analysis, legal document summarization, regulatory compliance and legal argumentation, where the temporal aspect is crucial in modeling.

\section*{Limitations}
While this study provides valuable insights for addressing temporal challenges in legal multi-label text classification tasks, it is important to acknowledge certain limitations. 

Firstly, the experiments and findings are based on six multi-label legal classification datasets. The extent to which the proposed approaches generalize across different legal domains or other text classification tasks remains to be explored. Further, the paper assumes the homogeneity of data within each time split, and the effectiveness of the proposed incremental training paradigm may vary based on the degree of diversity and complexity within each temporal split. 
Furthermore, while the streaming evaluation protocol is advocated for its reliability, its application to a broader range of datasets and tasks might introduce additional computational overhead. Investigating the protocol's feasibility and performance in diverse settings is essential for establishing its practicality. Lastly, the paper addresses temporal drift in legal tasks, but a more detailed exploration of the specific causes and characteristics of temporal drift within legal domains could enhance the understanding of the challenges involved.  

While the paper recognizes the existence of temporal drift, delving deeper into the specific characteristics of temporal shifts in legal text data (e.g., gradual vs. abrupt shifts) and their impact on model performance could contribute to more nuanced insights into the temporal dynamics at play. Unless one is explicitly examining the effect of a particular watershed event on the legal system and is informed by substantial relevant legal expertise, attributing large legal datasets to underlying factors accounting to inherent concept drift may turn out to be infeasible and we reserve that as a potential avenue for future work.  For instance, in ECtHR jurisprudence, one would need to find multiple near-identical cases with different outcomes. Finding those is hard and, if they were found, they would be of great interest to legal scholarship.

Additionally, this work assumes a pre-fixed static splits, following a standard deployment scenario, where the model is updated based on frequency. In prior work on the US Supreme Court \cite{katz2017general}, for example, a change in the court’s presiding judge marks natural boundaries at the supra-year level. However, such segmentation requires in-depth knowledge of the domain and collection. The challenge of finding an appropriate time split and can we have it determined dynamically, is difficult and would possibly require a qualitative exploration of how identifiable shifts in the data coincide with an expert’s intuition about the natural temporal segmentation of the collection.

\section*{Ethics Statement}
The legal text classification datasets used in this study are sourced from publicly available repositories and produced by previous studies. Though the judgment corpus from ECHR is not anonymized and contains the real names of the involved parties, we do not foresee any harm incurred by our experiments. The intention behind this research is to provide a nuanced understanding of the temporal challenges inherent in legal text classification tasks. The findings are intended to benefit legal practitioners, researchers, and organizations involved in legal document management by enhancing the adaptability and performance of classification models in the dynamic legal landscape. 

\bibliography{custom}
\bibliographystyle{acl_natbib}

\appendix

\section{Implementation Details}
\label{app-implementation}
We will release our code upon acceptance.

\noindent \textbf{Baseline Models} We fine-tune the models on our six datasets using binary cross-entropy loss and they are trained end-to-end using AdamW optimizer \cite{loshchilov2017decoupled}, learning rate of 2e-5, momentum of 0.9, and weight decay of 0.01. For ECHR tasks, to account for longer input length, we employ hierarchical models which can process 64 paragraphs each with 128 tokens. We use a batch size of 60 for UKLEX and EURLEX tasks, and 6 for ECHR tasks due to memory constraints. The models are trained for up to 20 epochs using early stopping with a patience of 3 on macro-F1 score of the validation set to prevent the model. We use 16-bit automatic mixed precision and gradient accumulation to accelerate training and save memory. All the experiments were performed on a GPU cluster with NVIDIA A40 48GB PCIe 4.0. 


\noindent \textbf{IFT models} At each time stamp $t$, we initialize the model with the best-performing model from the previous timestamp ($t-1$). A warm-up phase of 3 epochs is employed before applying the patience threshold. This initial warm-up phase allows the model to interact with the data in the current timestamp, providing ample time to discern drifts and adapt accordingly. The remaining hyperparameters are kept consistent with the baseline models, and the best model at each time stamp is selected based on the macro-F1 score on the validation set.


\noindent \textbf{Evaluation Metrics:} For UKLEX and EURLEX, we employ the standard approach to compute macro-F1, micro-F1, and m-RP scores on the list of labels. However, for ECHR, an additional label is introduced during inference to account for instances that do not violate (allege) any article. In Task A (B), where there are 14 (17) articles, this extra label is included in both targets and predictions and its value is set to 1 if none of the labels are 1 (indicating that all the labels are 0), following the methodology outlined by \citet{chalkidis2022lexglue}.

We use Wild-Time library \cite{yao2022wild} for implementing continual learning algorithms such as EWC, ER, AGEM, and temporal-invariant methods such as GroupDRO, DeepCORAL, and IRM. We use  AdapterHub framework \cite{pfeiffer2020AdapterHub} to implement LoRA and Adapters. 

\subsection{Continual Learning methods specific Hyper-parameters}
\noindent \textbf{EWC} \cite{kirkpatrick2017overcoming} We use the online version of EWC training to avoid memory overflow in the computation of Fischer information matrices, as detailed in \cite{kirkpatrick2017overcoming}. We set $\lambda$ of 0.5 which controls the strength of regularization-based EWC loss with a decay term for older data $\gamma$ set to default of 1.0. 

\noindent \textbf{ER} \cite{rolnick2019experience} We replay a mini-batch of data by random sampling from an evolving past data module after every 10 training steps for UKLEX and EURLEX tasks, and every 30 training steps for ECHR tasks.

\noindent \textbf{AGEM} \cite{lopez2017gradient} stores parameter gradients of data from past timestamps and uses reservoir strategy to sample the past to add to the memory buffer. We use a maximum size of 1000 for the memory buffer for rehearsal.

\noindent \textbf{LoRA} \cite{hu2021lora} We set the low-dimensional rank $r$ of the decomposition matrix to 8 and the $\alpha$ which accounts for scaling the reparameterization to 16.

\noindent \textbf{Adapters} \cite{houlsby2019parameter} We used bottleneck adapters with a reduction factor of 16 which controls the ratio between the hidden dimension and the bottleneck dimension. 

\subsection{Temporal Invariant learning methods specific Hyper-parameters}
For all the temporal invariant methods, we sample uniformly from each time period in the window, when treated as a single domain to apply domain invariant approaches to boundary-less temporal setting. We employ sliding windows of length 5. We set the number of domains to be considered to be 3. 

\noindent \textbf{DeepCORAL} \cite{sun2016deep} We found the model is highly sensitive to $\lambda$ which controls the penalty on alignment loss due to the challenging nature of its direct effect on feature representations. We performed an exhaustive search over the range and found 0.001 to work well in our setting. We compute alignment penalties between features from all pairs of domains sampled. 

\noindent \textbf{IRM} \cite{arjovsky2019invariant} We set IRM penalty factor $\lambda$ to be constant at 1.0. 

\noindent \textbf{GroupDRO} \cite{sagawa2019distributionally} Each instance in the minibatch is sampled with uniform probabilities from the entire domain.

\section{Evaluation}
\label{app-eval}

\subsection{Eval-Fix}
\begin{figure*}
    \centering
    \subfigure[UKLEX(S)]   {\includegraphics[width=\textwidth]{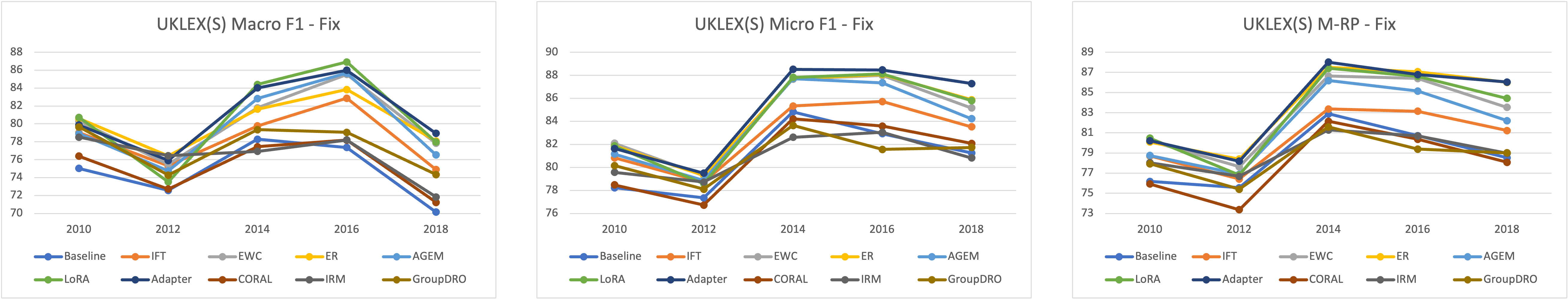}}
    
    \qquad
    \subfigure[UKLEX(M)]    {\includegraphics[width=\textwidth]{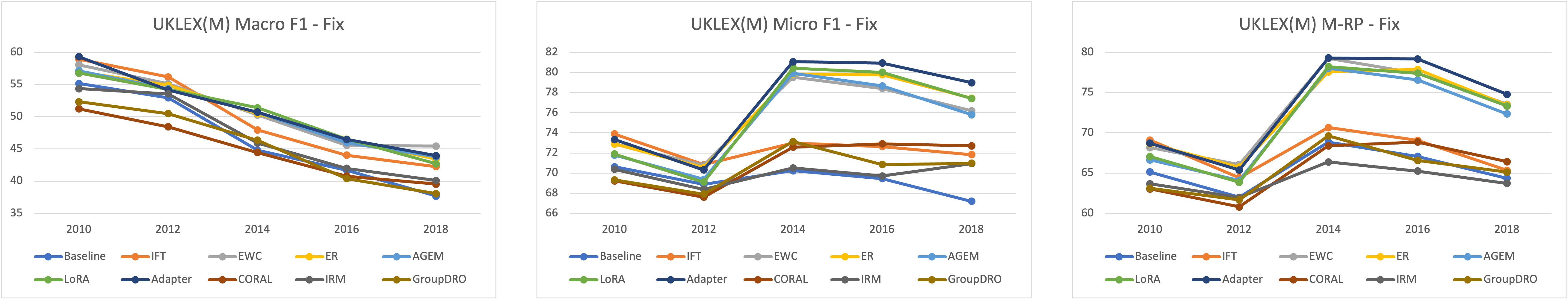}}

    \qquad
    \subfigure[EURLEX(S)]    {\includegraphics[width=\textwidth]{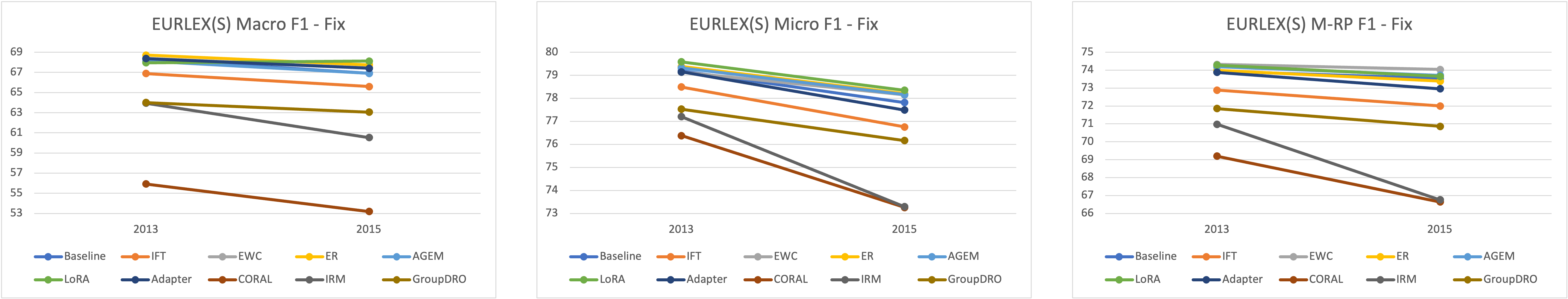}}

    \qquad
    \subfigure[EURLEX(M)]    {\includegraphics[width=\textwidth]{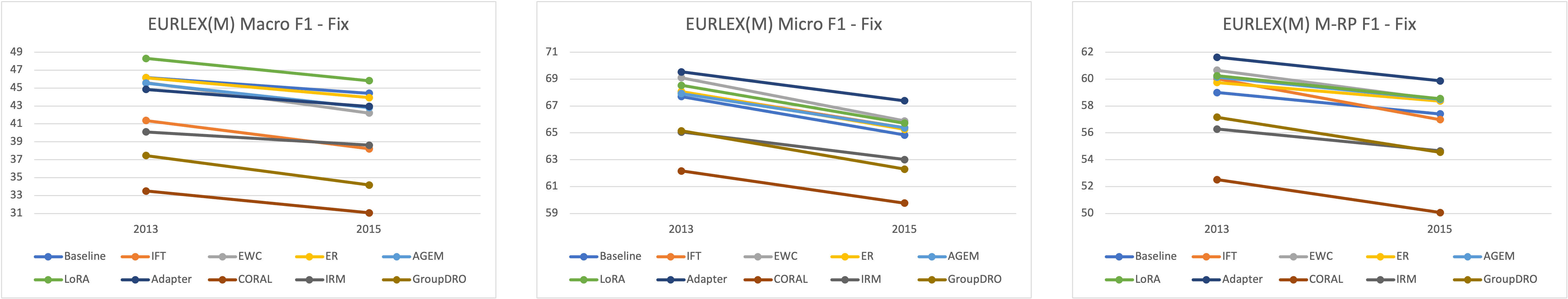}}

    \qquad
    \subfigure[ECHR(A)]    {\includegraphics[width=\textwidth]{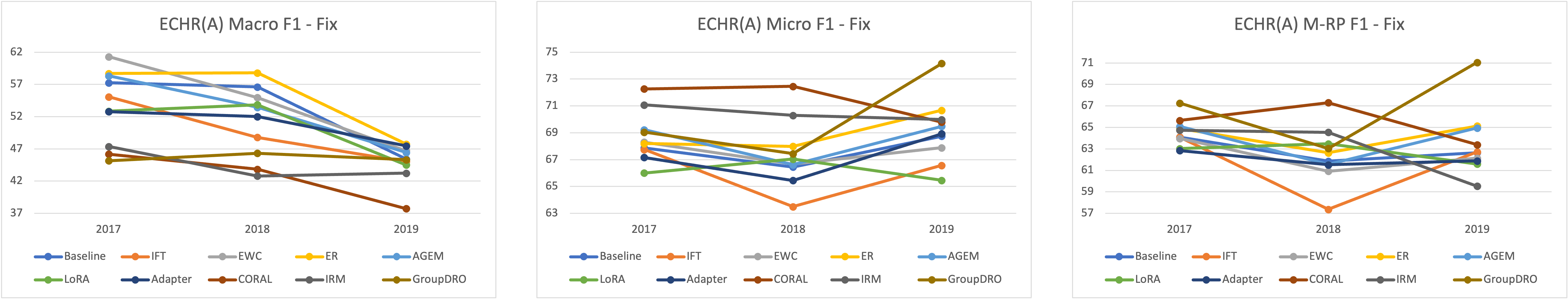}}

    \qquad
    \subfigure[ECHR(B)]    {\includegraphics[width=\textwidth]{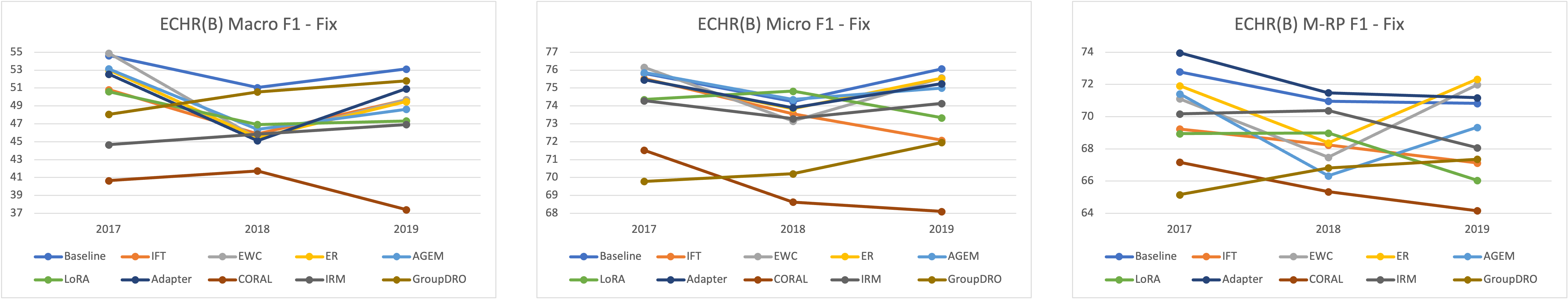}}

    \caption{Results on Eval-Fix Setting over six datasets.}
    \label{fig:evalfix}
\end{figure*}

We present the performance of each method for every time period within the fixed test split of the Eval-Fix settings across all datasets in Figure \ref{fig:evalfix}. Notably, we observe a consistent improvement of methods across all timestamps in the test split, particularly evident in the EURLEX(S,M) and UKLEX(S,M) datasets where the line plots corresponding to each method appear parallel over the years. However, this trend is not uniformly observed in the ECHR(A,B) datasets. We posit that this deviation may be attributed to the extreme sparsity of the dataset, causing different methods to excel on different subsets, thereby influencing performance based on the distribution of labels in that time stamp. While a declining trend over the years has been observed in most datasets, it is expected due to the distribution shift between the training and testing sets as the temporal distance increases with the test year. However, some datasets, like UKLEX(S), exhibit a 'V'-shaped trend. This observation prompts future investigations to discern the reasons behind such trends and to study the underlying shift or drift phenomena contributing to this unique pattern. Unraveling these intricacies can serve as motivation for future research to devise shift-specific mitigation strategies tailored to address the nuanced temporal dynamics present in these datasets.

\subsection{Eval-Stream}
\begin{figure*}
    \centering
    \subfigure[UKLEX(S)]   {\includegraphics[width=\textwidth]{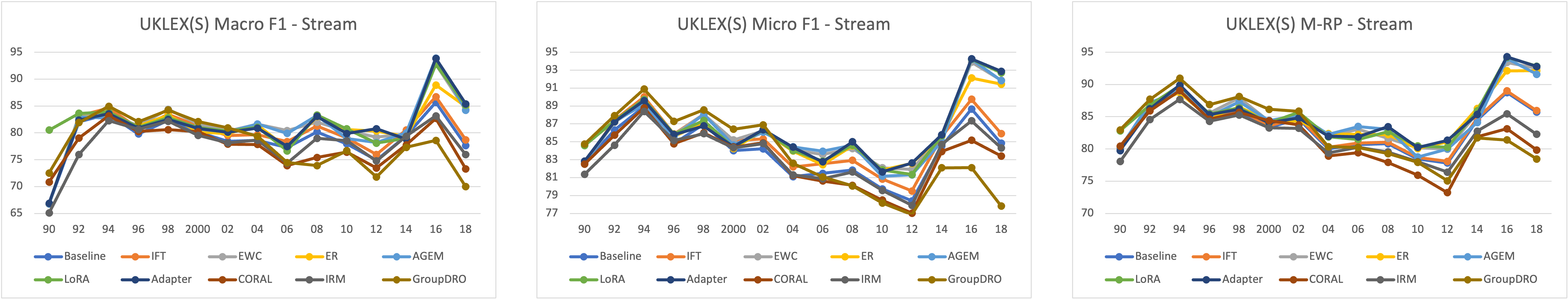}}
    
    \qquad
    \subfigure[UKLEX(M)]    {\includegraphics[width=\textwidth]{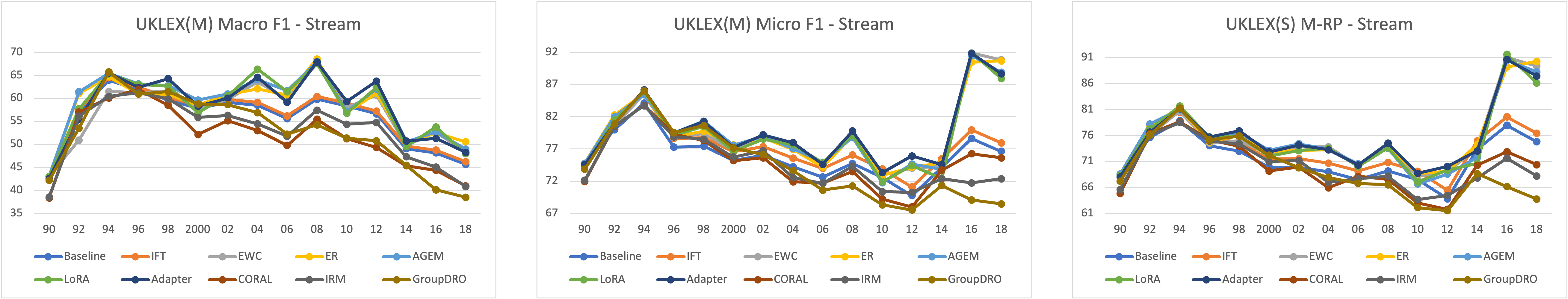}}

    \qquad
    \subfigure[EURLEX(S)]    {\includegraphics[width=\textwidth]{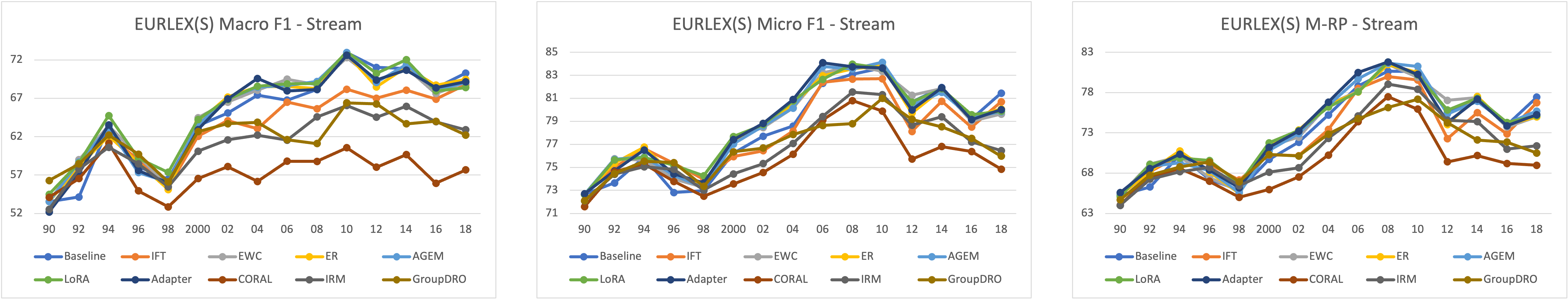}}

    \qquad
    \subfigure[EURLEX(M)]    {\includegraphics[width=\textwidth]{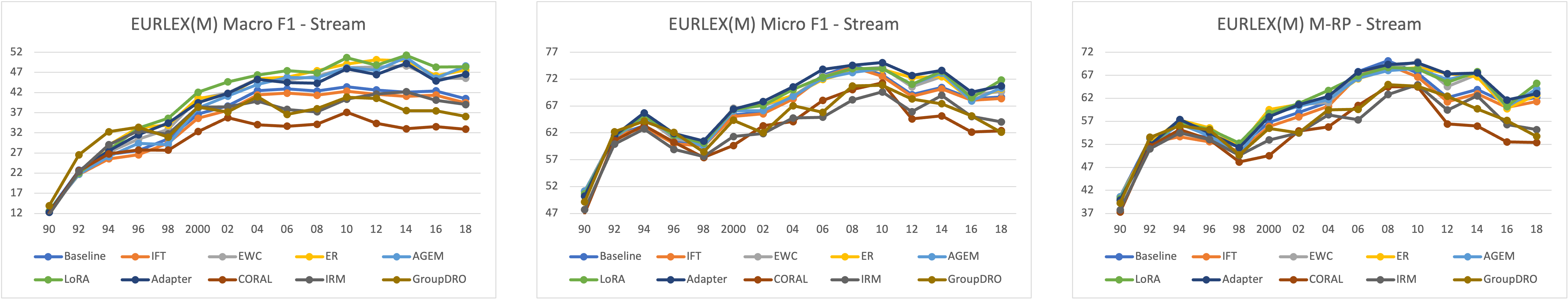}}

    \qquad
    \subfigure[ECHR(A)]    {\includegraphics[width=\textwidth]{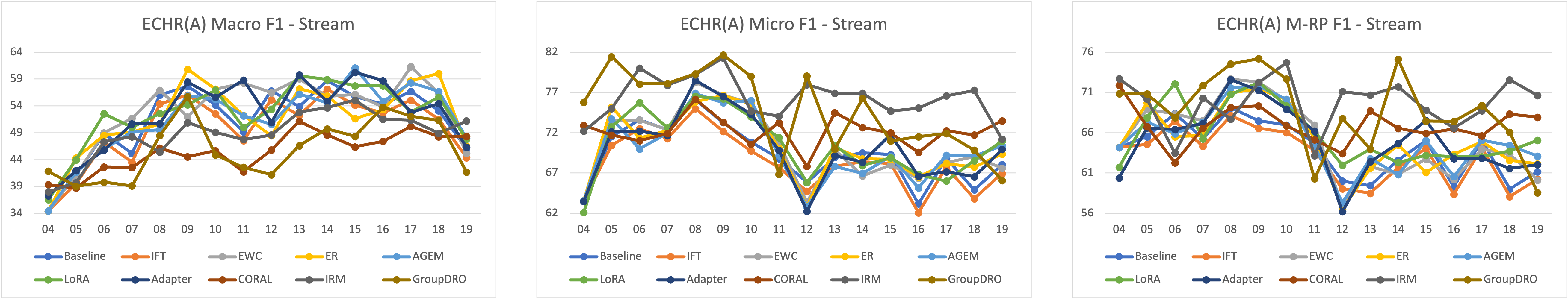}}

    \qquad
    \subfigure[ECHR(B)]    {\includegraphics[width=\textwidth]{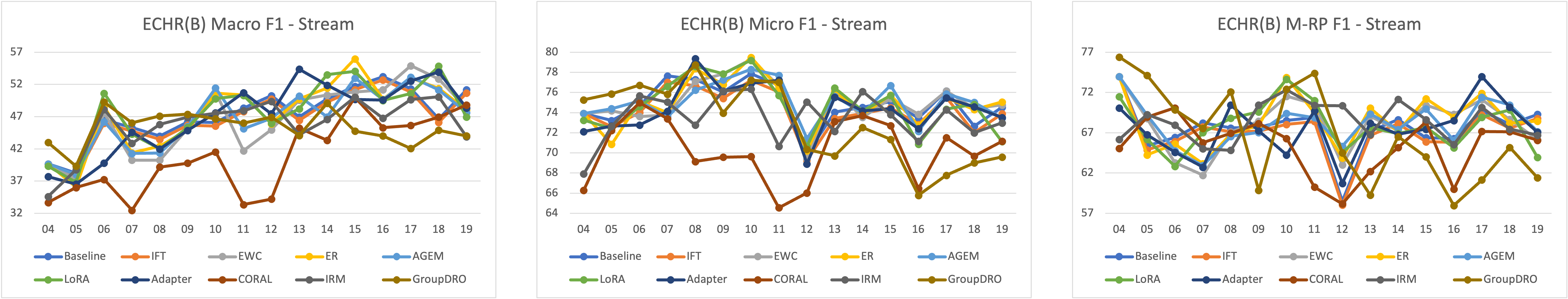}}

    \caption{Results on Eval-Stream Setting over six datasets.}
    \label{fig:evalsstream}
\end{figure*}

We evaluate the models on multiple splits and present the corresponding evaluation scores for subsequent time period at each split in Figure \ref{fig:evalsstream}. The consistent trend of uniform improvements across methods on UKLEX(S,M) and EURLEX(S,M), as well as non-uniform improvements on ECHR(A,B), is mirrored in the streaming setting as well. This non-uniform nature can prompt interesting case studies to decipher what models might be learning and leverage this information to design strategies for improving legal text classification models.

\end{document}